\documentclass[letterpaper, 10 pt, conference]{ieeeconf}  % Comment this line out if you need a4paper
\IEEEoverridecommandlockouts                              % This command is only needed if
\usepackage{graphics} % for pdf, bitmapped graphics files
\usepackage{times} % assumes new font selection scheme installed
\usepackage{amsmath} % assumes amsmath package installed
\usepackage{amssymb}  % assumes amsmath package installed
\usepackage{graphicx}
\usepackage{algorithm}
\usepackage{multirow}
\usepackage[noend]{algpseudocode}
\usepackage{url}
\usepackage{hyperref}
\usepackage[font=small]{caption}
\usepackage{subcaption}

\let\labelindent\relax
\usepackage{enumitem}
\setlist[enumerate]{leftmargin=*}
\setlist[itemize]{leftmargin=*}

\usepackage{color}

\usepackage[
    style=numeric,%ieee, 
    sorting=none,        % Bibliography is ordered by first appearance in the text
    sortcites=true,      % Sorts numbers numerically INSIDE the \cite{} brackets
    backend=biber, 
    maxbibnames=99,      % Forces all authors to be listed in the bibliography
    natbib=true,
    doi=false,   % Excludes DOI fields
    url=false,   % Excludes URL fields
    isbn=false,  % Optional: Excludes ISBN (common for IEEE to save space)
    eprint=false, % Optional: Excludes eprint/arXiv links    
    giveninits=true
]{biblatex}

\def\secref#1{Sec.~\ref{#1}}
\def\figref#1{Fig.~\ref{#1}}
\def\tabref#1{Tab.~\ref{#1}}
\def\eqref#1{Eq.~(\ref{#1})}

\title{\LARGE \bf Autonomous Aerial Non-Destructive Testing:  Ultrasound Inspection with a Commercial Quadrotor in an Unstructured Environment}

\author{Ruben Veenstra$^{1}$, Barbara Bazzana$^{1}$, Sander Smits$^{1}$, and Antonio Franchi$^{1,2}$% <-this % stops a space
\thanks{$^{1}$ Robotics and Mechatronics lab, Faculty of Electrical Engineering, Mathematics \& Computer Science, University of Twente, Enschede, The Netherlands. \href{mailto:r.veenstra@utwente.nl}{r.veenstra@utwente.nl}, \href{mailto:b.bazzana@utwente.nl}{b.bazzana@utwente.nl}, \href{mailto:s.smits@utwente.nl}{s.smits@utwente.nl}, \href{mailto:schol@r-franchi.eu}{schol@r-franchi.eu}}%
\thanks{$^{2}$Department of Computer, Control and Management Engineering, Sapienza University of Rome, 00185 Rome, Italy.}%
\thanks{This work was partially funded by the Horizon Europe research agreement no. 101120732 (AUTOASSESS)}%
}

\begin{document}

\maketitle
\thispagestyle{empty}
\pagestyle{empty}

%%%%%%%%%%%%%%%%%%%%%%%%%%%%%%%%%%%%%%%%%%%%%%%%%%%%%%%%%%%%%%%%%%%%%%%%%%%%%%%%
    
    \begin{abstract}
This work presents an integrated control and software architecture that enables arguably the first fully autonomous, contact-based non-destructive testing (NDT) using a commercial multirotor originally restricted to remotely-piloted operations. To allow autonomous operation with an off-the-shelf platform, we developed a real-time framework that interfaces directly with its onboard sensor suite. The architecture features a multi-rate control scheme: low-level control is executed at 200 Hz, force estimation at 100 Hz, while an admittance filter and trajectory planner operate at 50 Hz, ultimately supplying acceleration and yaw rate commands to the internal flight controller. We validate the system through physical experiments on a Flyability Elios 3 quadrotor equipped with an ultrasound payload. Relying exclusively on onboard sensing, the vehicle successfully performs autonomous NDT measurements within an unstructured, industrial-like environment. This work demonstrates the viability of retrofitting off-the-shelf platforms for autonomous physical interaction, paving the way for safe, contact-based inspection of hazardous and confined infrastructure.
\end{abstract}

%%%%%%%%%%%%%%%%%%%%%%%%%%%%%%%%%%%%%%%%%%%%%%%%%%%%%%%%%%%%%%%%%%%%%%%%%%%%%%%%
\section{Introduction}
\label{sec:intro}

Critical infrastructures, such as ballast tanks, cargo hulls, power and oil plants, require inspection and monitoring in harsh and difficult-to-access environments. A very common inspection paradigm is contact-based Non-Destructive Testing (NDT), which allows for safe, efficient, and accurate inspections. As inspection requires access to risky and unhealthy environments, the need for robotic inspection has progressively expanded, in parallel with that for advanced sensor technologies that can be integrated on such platforms. Among them, aerial robots are particularly relevant, thanks to their 3D workspace, which allows them to reach otherwise inaccessible areas, such as confined or at-height spaces.

Motivated by this need, research has focused on developing (i) control schemes for compliant interaction, leveraging the literature on fixed-based manipulators; (ii) custom HW designs of aerial platforms or of end-effectors, where an increased maneuverability achieved through additional inputs or kinematic DOFs would allow a more dexterous flight~\cite{2021f-HamUsaSabStaTogFra,2021g-OllTogSuaLeeFra}. Despite the advantages of the enhanced maneuverability of redundant platforms, the market leader OEMs (Original Equipment Manufacturers)\footnote{\url{https://www.flyability.com/}}\footnote{\url{https://www.scoutdi.com/}}  for confined-space inspection still extensively use quadrotor platforms, due to their versatility, energy efficiency, and compactness. In such cases, the reduced control authority is compensated through HW design, using mechanically compliant end-effectors with magnets at the tip.
\begin{figure}[t]
  \centering
\includegraphics[width=0.99\columnwidth]{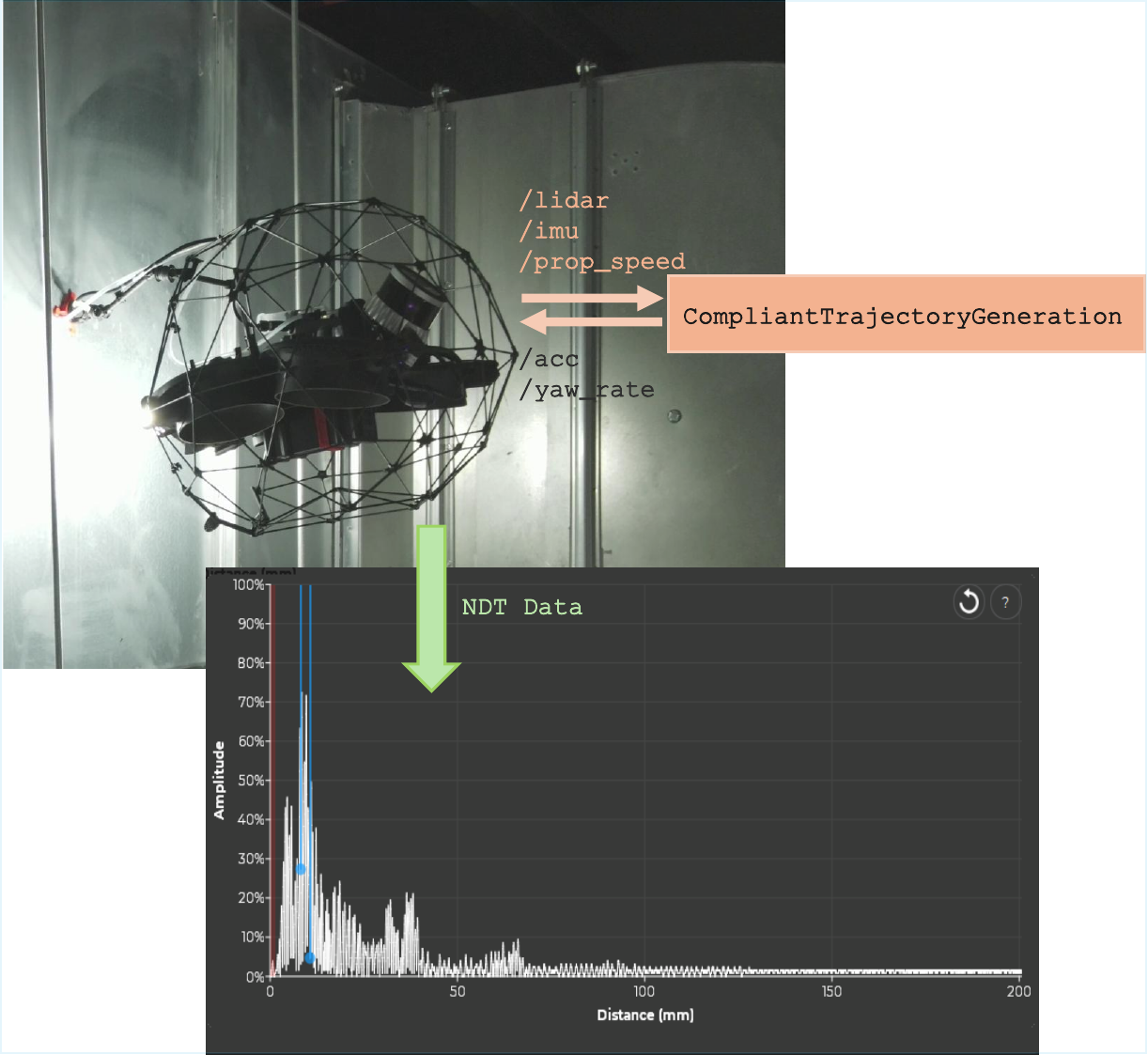}
  \caption{The overview of our system, allowing to get NDT data through compliant interaction with an inspection surface, using onboard sensors. The algorithm is tested on the Flyability Elios3 in the industrial testbed at the University of Twente, Enschede, The Netherlands.}
  \label{fig:motivation}
\end{figure}
In this work, we present the first autonomous NDT data collection system developed for the Flyability Elios~3, a commercial underactuated quadrotor equipped with onboard localization and an Ultrasonic Testing (UT) payload for thickness measurements. Specifically, we detail: (i)~the state machine governing the autonomous inspection, (ii)~the core control modules enabling stable physical interaction, and (iii)~the system architecture that integrates these components relying exclusively on the onboard sensor suite. Like many off-the-shelf platforms, the Elios~3 is fundamentally designed for remotely piloted operations, particularly during the highly demanding physical contact phase. Endowing it with autonomous physical inspection capabilities allows us to leverage the robust safety measures and established ecosystem of a mature commercial platform, all while keeping human operators entirely out of hazardous environments. The platform executing an autonomous inspection in an industrial-like testbed is depicted in~\figref{fig:motivation}. 

In addition to describing the system overview, as applied from the theory of aerial physical interaction, we provide practical insights into the guiding factors for NDT contact-based inspection, starting from the UT post-processed data, which we have access to on the targeted commercial platform. Finally, we show the validity of the proposed approach through real-world experiments in an unstructured industrial testbed at the University of Twente, Enschede, The Netherlands.

Summarizing, the contribution of this work goes as follows: (i) an overview of a physical interaction paradigm for a commercial quadrotor specialized for autonumous inspection in confined spaces; (ii) a description of the guiding factors for NDT contact-based inspection; (iii) an experimental validation including UT measurement data for further reference.

%%%%%%%%%%%%%%%%%%%%%%%%%%%%%%%%%%%%%%%%%%%%%%%%%%%%%%%%%%%%%%%%%%%%%%%%%%%%%%%%
\section{Related Work}
\label{sec:related}
Motivated by the importance  of autonomous inspections, research has recently focused on developing NDT-oriented aerial robotic platforms applying advances in the control of aerial platforms for physical interaction~\cite{2021g-OllTogSuaLeeFra, 2019h-RylMusPieCatAntCacBicFra}. 
Tognon \emph{et al.}~\cite{2019e-TogTelGasSabBicMalLanSanRevCorFra} describe a truly-redundant aerial manipulator, composed of a 6 DOFs hexarotor endowed with a 2DOFs lightweight arm carrying an Eddy current probe for thickness measurement, while sliding on a metallic pipe. The system relies on a Motion Capture system for localization. 

Further research on control of redundant aerial platforms for contact-based inspection has focused on exploiting the additional DOFs to (i) integrate visual feedback to adapt to different contact geometries~\cite{rashad2020iros}, (ii) employ selective impedance control and parallel force-motion strategies~\cite{bodie2019omnidirectional}, and (iii) optimization-based control to satisfy various control objectives during flight~\cite{peric2021icra, 2020a-NavSabTogPucFra}. 

In parallel with the control strategy, on-the-field deployment of the proposed drone technology requires (i) integration of onboard sensors for localization~\cite{kulkarni2025icar} and (ii) combination with vision-based algorithms to detect the points of interest~\cite{brogaard2024science}. Sanchez-Cuevas \emph{et al.}~\cite{sanchez2020sensors} equip a fully-actuated hexarotor with a 3DOFs lightweight arm endowed with a sensorized passive joint for force regulation, and onboard sensors for localization in GNSS-denied environments. Lippiello~\emph{et al.}~\cite{lippiello2016ral} propose a hybrid visual-servoing approach to leverage visual inputs in the control pipeline.

Finally, the design and equipment of the end-effectors ultimately allow the collection of reliable NDT measurements~\cite{suarez2016iros}.  D'Angelo~\emph{et al.}~\cite{dangelo2025drones} propose a fully-customized aerial manipulator composed of a tilting quadrotor and an articulated robotic arm, equipped with a UT payload for thickness measurements, an ATI force sensor used by the force/impedance control scheme, and an onboard suite for localization encompassing ToF sensors, a Unilidar L1 PM 3D LiDAR, a Realsense T265, an ArkFlow optical flow sensor, and the internal sensors of the Pixhawk controller. Trujillo \emph{et al.}~\cite{trujillo2019sensors} introduce a novel aerial robotic manipulator, consisting of an eight-tilted rotor configuration and an end-effector with wheels, that provides GNSS-free physical contact inspection, using on-board sensors to maintain contact with the inspection surface, able to perform A-scan, B-scan, or C-scan inspections.  Thurnherr \emph{et al.}~\cite{thurnherr2025elsevier} show a novel payload tested on the Voliro platform, allowing for thickness measurements of concrete structures.   

Our work aligns with the recent literature in NDT using UAV~\cite{bi2025advances}, which can be categorized into: crawler systems such as the one proposed by~\cite{faclonscan2024}, customized versions of commercially available UAVs such as the AscTec Firefly~\cite{zhang2018ius}, and off-the-shelf commercial UAVs, such as the Voliro platform used by~\cite{watson2022trase}. Leveraging the experience gained on custom platforms, we here propose the first integration of physical interaction control schemes with off-the-shelf NDT-ready commercial quadrotors, designed for operation in confined and unstructured GNSS-denied environments, and test it on the Flyability Elios3 drone for confined-space inspection.

%%%%%%%%%%%%%%%%%%%%%%%%%%%%%%%%%%%%%%%%%%%%%%%%%%%%%%%%%%%%%%%%%%%%%%%%%%%%%%%%
\section{System Overview}

% Flowchart instead?
\label{sec:system-overview}

\subsection{Hardware description}

\begin{figure}[t]
  \centering
 \includegraphics[width=0.99\columnwidth]{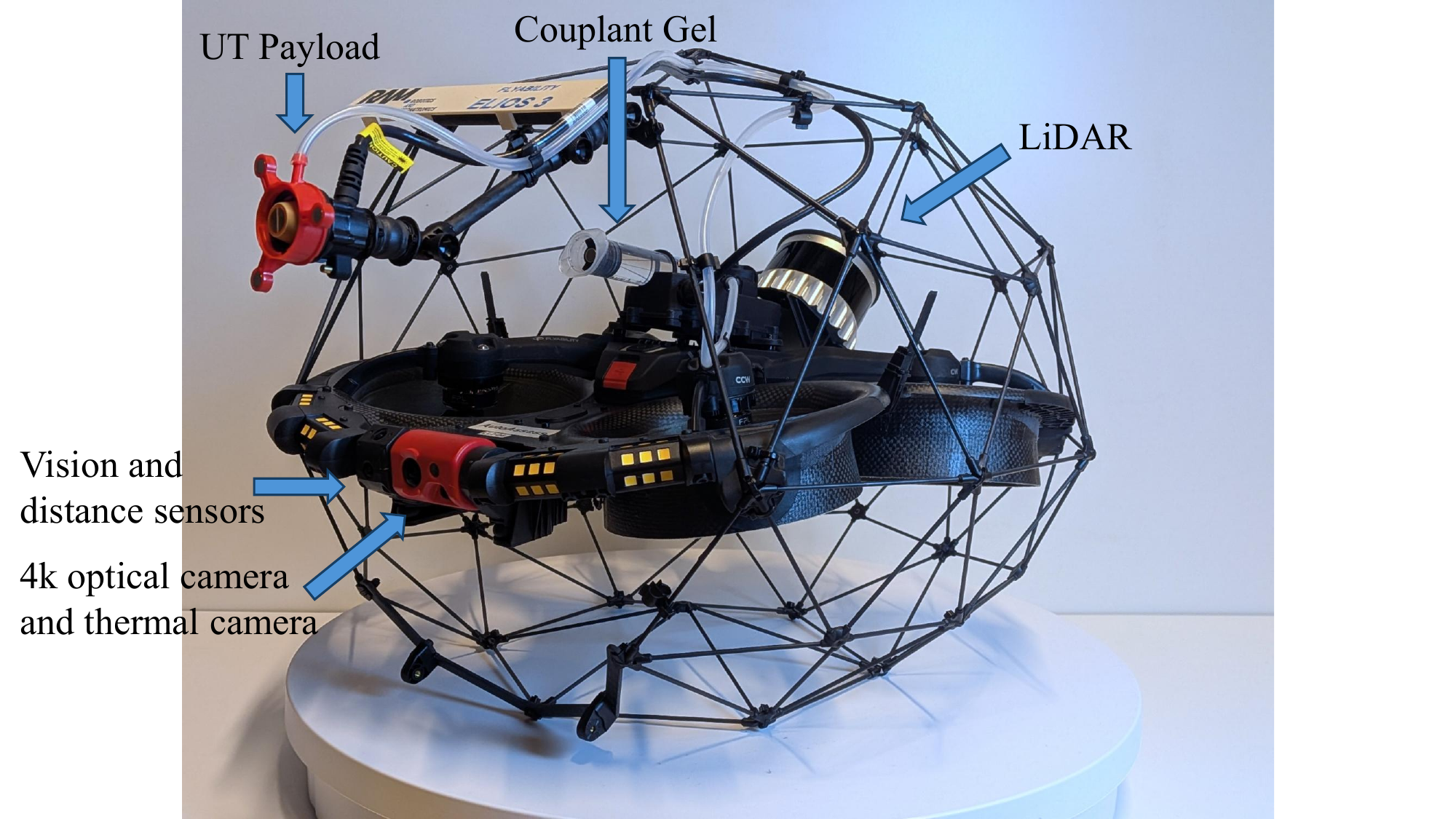}
  \caption{Flyability Elios3 System with highlight of the onboard sensors and equipment. LiDAR, vision and distance sensors are used for localization, the UT payload and the couplant gel for NDT data collection, the 4k optical camera and the thermal camera are shown online to the operator to provide with a clear understanding of the scenario, when operating the drone remotely.}
  \label{fig:elios}
\end{figure}

The Flyability Elios 3, shown in ~\figref{fig:elios}, is a commercial unmanned aerial system (UAS) specifically designed for inspection tasks in confined and hazardous environments. The vehicle is enclosed within a protective collision-tolerant cage, enabling safe contact with surrounding structures and allowing operation in cluttered or GPS-denied spaces where conventional multirotor platforms would be at risk of damage.

The platform is equipped with optical and LiDAR sensors, which enable simultaneous localization and mapping (SLAM) in confined and global navigation satellite system (GNSS)-denied environments. This capability allows operators to generate accurate three-dimensional maps while obtaining precise state estimation, required for stabilization.

In addition to visual and LiDAR-based inspection capabilities, the system can be extended with an ultrasonic testing (UT) payload, enabling Non-Destructive Testing (NDT) physical inspection of structures. In particular, the system supports A-scan ultrasonic measurements for wall thickness assessment and defect detection. The UT payload consists of an ultrasonic probe mounted on the protective cage, with a manually adjustable mechanism that allows the probe to be oriented relative to surfaces of varying inclinations and geometries. Further, the mechanism is compliant to facilitate collision tolerance and some degree of decoupling between the UAS and environment.

To ensure reliable measurements, a magnetic hood is positioned over the probe to maintain stable contact with ferromagnetic surfaces. Proper probe-to-surface contact is critical for measurement accuracy, as air gaps significantly degrade ultrasonic signal transmission. Once contact has been established, ultrasonic couplant can be dispensed to improve acoustic coupling between the probe and the inspection surface, thereby enhancing signal quality and measurement accuracy.

\subsection{Inspection algorithm}\label{sec:algo}

\begin{figure}[t]
  \centering
 \includegraphics[width=\columnwidth]{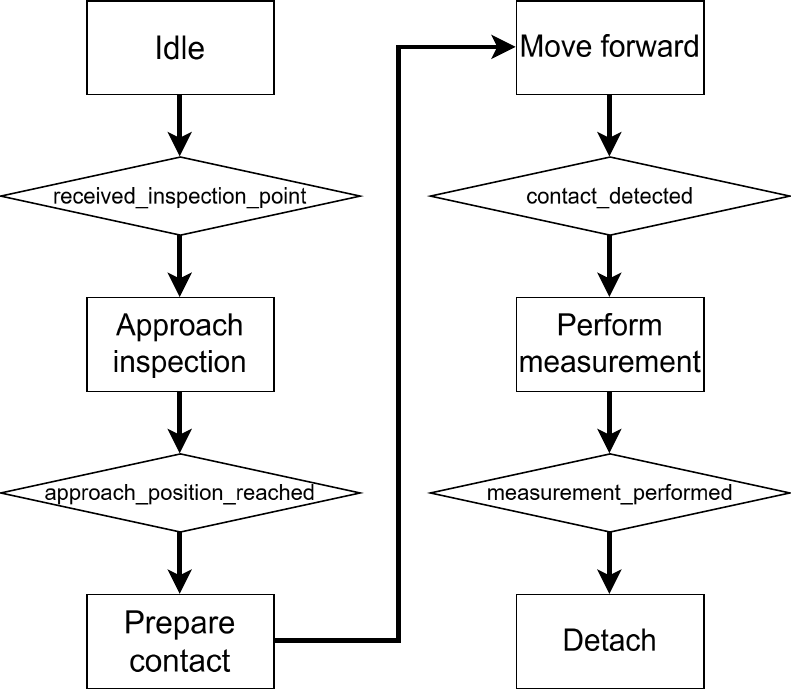}
  \caption{Inspection algorithm flow diagram.}
  \label{fig:flow}
\end{figure}

%\begin{algorithm} \caption{NDT contact-based inspection} \label{algo:ndt-inspection} \
%\begin{algorithmic} 
%\State \texttt{followApproachTrajectory()}
%\While{\texttt{inspect}} \State \texttt{followCompliantTrajectory()} \EndWhile 
%\State \texttt{followDetachTrajectory()}
%\end{algorithmic} 
%\end{algorithm}

To enable autonomous and safe physical inspection, the controller must ensure that interaction forces remain bounded. Accurate inspection of target points requires precise trajectory tracking, which in turn demands sufficiently high controller stiffness. However, during contact, high stiffness can cause rapid actuator saturation. To balance these conflicting requirements, force control strategies have been developed in robotics. In this work, we employ an indirect force control strategy, admittance control, which decouples motion control from interaction behavior. This approach enables accurate trajectory tracking while maintaining compliance with external forces~\cite{Villani2008}. Overall, the inspection algorithm consists of three main components:
\begin{itemize}
    \item a trajectory generation module, which uses the open-source CPP implementation of Richter~\emph{et al.}~\cite{richter2016polynomial} (\url{https://github.com/ethz-asl/mav_trajectory_generation}), to generate smooth 50Hz trajectories given sets of waypoints;
    \item an admittance filter module, which, given as input a reference smooth trajectory, computes compliant trajectories for stable physical interaction, using the online estimate of the external force
    \item the acceleration-based force observer module, which provides an estimate of the external force on the system, required for the admittance filter and contact detection.  
\end{itemize}

The inspection algorithm is designed as a modular component that can be integrated into an autonomous software stack. Given a map of the asset to be inspected, potentially obtained during an initial exploration phase, inspection points can be identified either manually or through machine-learning-based methods~\cite{andersen2023science}. A path planner then navigates the platform to the vicinity of the inspection point~\cite{bircher2015icra, macleod2018trase}, possibly enhanced using semantics~\cite{dharmadhikari2025trfr}.
Once the system is within range, the inspection algorithm is triggered by providing the pose (position and orientation) of the inspection point, which can be extracted from the map. The algorithm subsequently manages the physical interaction with the structure autonomously through a sequence of predefined steps. A flow diagram of the inspection algorithm is shown in ~\figref{fig:flow}. Each state is described in more detail as follows:
\begin{enumerate}
    \item \textbf{Idle:} The system waits for an inspection request.
    \item \textbf{Approach Inspection:} The quadrotor navigates to a pre-contact waypoint that aligns the probe with the surface while maintaining an offset of 0.5 m normal to the surface.
    \item \textbf{Prepare Contact:} The admittance controller is enabled, and biases in the force observer, possibly caused by aerodynamic effects in confined spaces, are estimated and subtracted.
    %\item \textbf{Move Forward:} The quadrotor advances slowly toward the surface until contact is detected. Contact is detected when the norm of the estimated external force exceeds 2~N {\color{red} AF: why 2N? is it a particular universal physical constant? try to separate the actual values you use in the implementation from the abstract quantities needed for the algorithm to work. Also it woudl be interesting to discuss how you arrived at this value, is it trial and error, is it jsut the just the first value which you tried?}.
    \item \textbf{Move Forward:} The quadrotor advances slowly toward the surface to establish contact. The required contact force, dependent on the sensor, can be selected by the user. Based on this value, the system computes the trajectory depth needed to achieve the desired force. Contact is considered detected when the norm of the estimated external force exceeds the user-defined threshold.
    \item \textbf{Perform Measurement:} Couplant is dispensed and the UT measurement is executed.
    \item \textbf{Detach:} After a successful measurement, a detachment maneuver is performed. The optimal detachment maneuver may depend on the sensor. For example, it was found that the magnetic hood for UT sensor of the Elios3 detaches most easily when applying a torque by performing a 60° yaw of the vehicle while moving backward and laterally.
\end{enumerate}

%The algorithm~\algref{algo:ndt-inspection} summarizes the three phases of a contact-based inspection mission, consisting of (1) approach, during which control tuning, smooth trajectory generation, and waypoint selection are crucial; (2) contact, where the admittance filter and wrench observer tuning play a major role, along with the HW design; and (3) detach, where the main criticality is the selection of suitable waypoints to overcome the force due to the magnets. In the following, we detail (i) the generation of compliant trajectories, (ii) the definition of suitable waypoints for the different phases of an inspection mission, and (iii) the KPIs for a good contact-based inspection.

\subsection{Control architecture}
\begin{figure*}[t]
  \centering
 \includegraphics[width=0.99\textwidth]{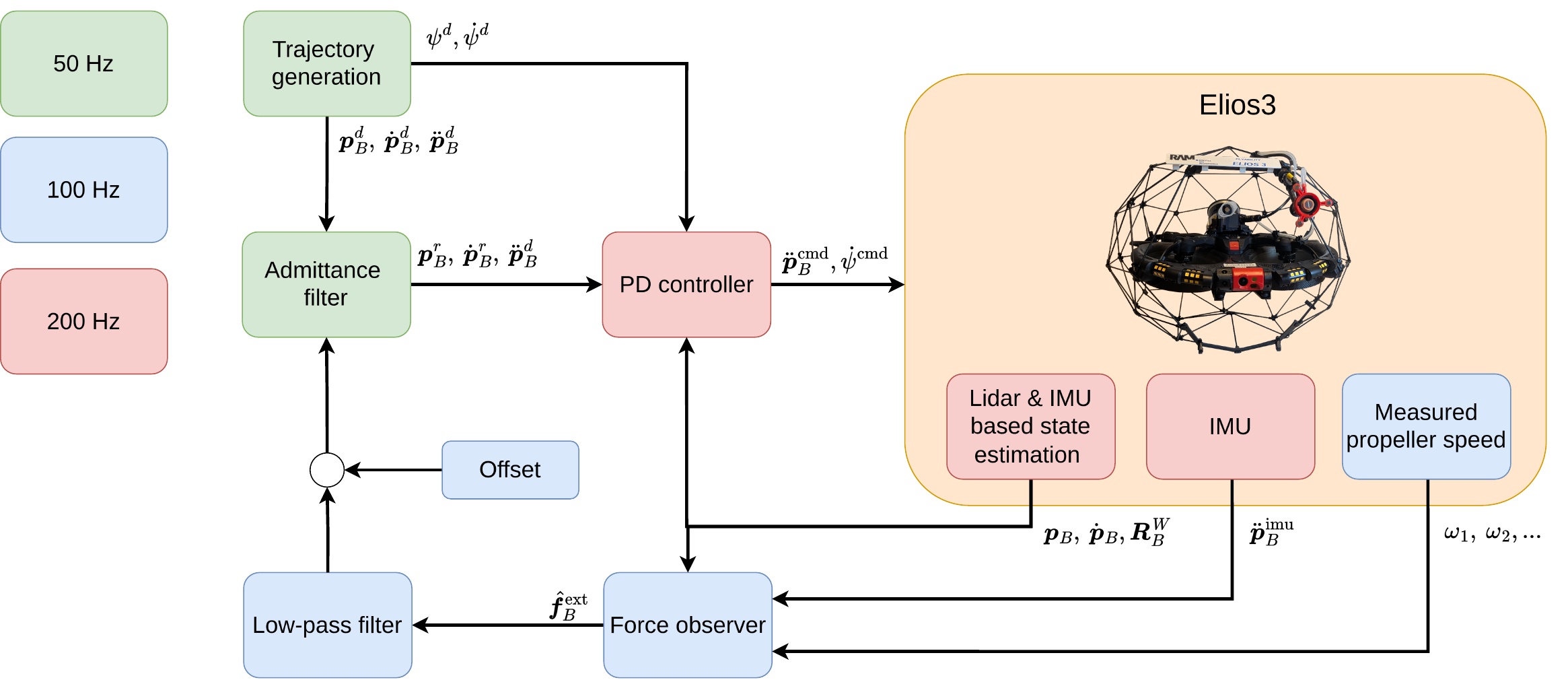}
  \caption{The constitutive modules of the control architecture, with the exchanged signals. The low-level control is proprietary software of Flyability, which is enclosed but can reliably follow acceleration and yaw rate setpoints.}
  \label{fig:overview}
\end{figure*}

In this section, the components of the physical interaction controller are described in detail. An overview of the framework is provided in ~\figref{fig:overview}. We start by presenting the translational dynamics, followed by the admittance filter, force observer, and finally the PD controller.

\paragraph{Translational dynamics}
For the modeling we define the following frames:
\begin{itemize}
    \item $\mathcal{F}_W$: The inertial world frame which is created by the system at startup. 
    \item $\mathcal{F}_B$: The body frame which is centered at the CoM of the vehicle. Oriented with Front-Left-Up.
    \item $\mathcal{F}_{P_i}$: The $i^\text{th}$ propeller frame attached at the center of the rotor with its z-axis aligned with the axis of rotation of the propeller.
\end{itemize}

The quadrotor, subject to an external force, can be modeled as a rigid body in free space. The translational dynamics of the quadrotor are described as:
\begin{equation} 
\label{eq:dynamics}
m \boldsymbol{\ddot{p}}_{B} = -mg\boldsymbol{e_3} + \boldsymbol{f}^{\text{rot}}_B + \boldsymbol{f}^{\text{ext}}_B
\end{equation}
where  $\boldsymbol{p}_{B} \in \mathbb{R}^{3}$ is the position of the robot in the inertial frame; $m$ is the system's mass; $g$ is the magnitude of acceleration due to gravity; $\boldsymbol{f}^{\text{ext}}_B$ is the external force applied at the body frame of the robot; $\boldsymbol{f}^{\text{rot}}_B$ is the actuation force applied to the body frame of the robot via the rotors. 

The thrust force generated by the $i^{\text{th}}$ propeller, expressed in the body frame, can be estimated by the quadratic model,  $^B\boldsymbol{f}_{P_i} = c_f |\omega_i | \omega_i \boldsymbol{z}_{P_i}$, where $c_f > 0$ is the thrust coefficient and $\omega_i$ is its rotational speed about the $\boldsymbol{z}_{P_i}$ axis~\cite{2021f-HamUsaSabStaTogFra}. The total actuation force in body frame is thus represented by $^B\boldsymbol{f}^{\text{rot}}_{B} = \sum_{i=1}^4 \:^B\boldsymbol{f}_{P_i}$. The corresponding  actuation force expressed in the inertial frame is, $\boldsymbol{f}^{\text{rot}}_{B} = \boldsymbol{R}^W_B \: ^B\boldsymbol{f}^{\text{rot}}_{B}$, where $\boldsymbol{R}^W_B  \in SO(3)$ denotes the rotation matrix from body frame to inertial world frame.

\paragraph{Force Estimator} 
An acceleration-based force observer is employed to obtain an estimate of the external force on the system, avoiding the need for direct measurement with an expensive and heavy force-torque sensor. The approach in~\cite{2019h-RylMusPieCatAntCacBicFra}, is followed here. Assuming a slow-varying external wrench, i.e.,~$\dot{{\boldsymbol{f}}}^{\text{ext}}_B \approx 0$, using model \eqref{eq:dynamics} to derive $\boldsymbol{f}^{\text{ext}}_B$ leveraging (i) the IMU measured acceleration to compute $\boldsymbol{\ddot{p}}_{B}$ and (ii) the measured propeller thrust to compute $\boldsymbol{f}^{\text{ext}}_B$, the estimated external force can be found by integrating the equation:

\begin{equation}
\label{eq:force_est}
\dot{\hat{\boldsymbol{f}}}^{\text{ext}}_B = -\boldsymbol{L} \: \hat{\boldsymbol{f}}^{\text{ext}}_B + \boldsymbol{L}(m\boldsymbol{\ddot{p}}_B + m g \boldsymbol{e}_3 - \boldsymbol{f}^{\text{rot}}_{B})
\end{equation}
where $\boldsymbol{L}>0 \in \mathbb{R}^{3\times3}$ is the force observer gain; $\hat{\ast}$ represents the estimated value of $\ast$. An additional first-order low-pass filter is applied to the signal to further attenuate high-frequency noise. Moreover, aerodynamic effects and inaccuracies in the measurement or estimation of parameters may introduce biases in the force estimate. Assuming these biases to be slow time-varying  they can be identified during flight and subsequently subtracted from the estimated force.

The thrust coefficient, $c_f$, needed for estimation of the thrust force produced, is identified through in-flight experiments. The mass of the platform is varied incrementally by attaching external masses up to 400 g, in 100 g increments, and recording the average measured propeller speeds during 10 second hovering periods. The dynamics of \eqref{eq:dynamics} simplify to:
\begin{equation} 
\label{eq:thrust_coef}
m_jg\boldsymbol{e_3} = \boldsymbol{f}^{\text{rot},j}_{B} = c_f \boldsymbol{R}^{W,j}_B \: \sum_{i=1}^4 \:|\omega_{i,j} | \omega_{i,j} \boldsymbol{z}_{P_i}
\end{equation}
The coefficient is subsequently estimated via a least-squares estimation (LSE) from the data pairs, $\left(m_j g \boldsymbol{e}_3, \boldsymbol{f}^{\text{rot},j}_{B}\right)$, which represent the weight force and the corresponding rotor-generated force for the $j$-th experiment.

\paragraph{Admittance Filter}
The admittance filter simulates a second-order linear dynamical system that maps external forces to motion. It has inputs (1) the estimated interaction force at the body, $\hat{\boldsymbol{f}}_B^{\text{ext}}$, and (2) the nominal desired trajectory from the trajectory generation, denoted with superscript $^d$, and as output the reference trajectory compliant with the environment, denoted with superscript $^r$:
\begin{equation} 
\label{eq:admittance}
\boldsymbol{M} \ddot{\boldsymbol{e}} + \boldsymbol{D} \dot{\boldsymbol{e}} + \boldsymbol{K}  \boldsymbol{e} = \hat{\boldsymbol{f}}_B^{\text{ext}}
\end{equation}
where $\boldsymbol{M},\boldsymbol{D},\boldsymbol{K}>0 \in \mathbb{R}^{3 \times 3}$ are the virtual mass, damping, and~stiffness matrices, respectively; $\boldsymbol{e} = \boldsymbol{p}^d_B - \boldsymbol{p}^r_B$, is the position error between the trajectory generated position $\boldsymbol{p}^d$, and compliant reference position $\boldsymbol{p}^r$. By appropriately selecting the virtual inertia, damping, and stiffness parameters, the filter shapes the system’s dynamic response, enabling compliant interaction while 
preserving stable and well-behaved motion.

\paragraph{PD Pose Controller}
The Elios3 accepts acceleration and yaw-rate control inputs. The objective of the PD controller is to drive the pose tracking error to zero. To accurately track position, velocity, acceleration, and yaw trajectories, the required acceleration and yaw rate commands are computed as follows:
\begin{equation}
\label{eq:PD}
\ddot{\boldsymbol{p}}^{\text{cmd}}_B = \ddot{\boldsymbol{p}}^r_B + \boldsymbol{K}_p(\boldsymbol{p}^r_B-\boldsymbol{p}_B)  +\boldsymbol{K}_v(\dot{\boldsymbol{p}}^r_B-\dot{\boldsymbol{p}}_B) + g\boldsymbol{e}_3-\frac{1}{m}\hat{\boldsymbol{f}}^{\text{ext}}_B
\end{equation}
\begin{equation}
\label{eq:yaw}
\dot{\psi}^{\text{cmd}} = \dot{\psi}^d + \boldsymbol{K}_{\psi}(\psi^d-\psi)  
\end{equation}
Where $\psi, \:\dot{\psi} \in \mathbb{R}$ are the yaw and yaw rate respectively; For positive-definite $\boldsymbol{K}_{\ast}\in \mathbb{R}^{3\times3}$ the linear error dynamics exponentially converge to zero. 

\subsection{Guiding Factors for NDT contact-based inspection}
We consider the inspection successful when a sequence of \textit{good and stable} readings has been obtained from the UT sensor. In our case, we have access to this information via the Flyability Cockpit App\footnote{\url{https://knowledge.flyability.com/aircraft/elios-3/elios-3-cockpit-app}}; if not available from the sensor producer, a filter can be implemented to estimate the stability of the measurement. Through testing, we have identified the following guiding factors for successful NDT data acquisition:
\begin{itemize}
    \item \textit{smooth} trajectory generation to ensure accurate tracking of a set of waypoints without abrupt maneuvers;
    \item \textit{compliant} physical interaction to reliably maintain contact, relying on an online estimate of the contact force;
    \item \textit{task-driven tuning} of the reference pushing force to trade-off measurement stability and energy consumption during physical interaction.
\end{itemize}
The physical interaction scheme proposed here is agnostic to the reference pushing force. As we did for our system, the user should tune the reference, depending on (1) the inspected surface and (2) the sensor used. In fact, the presence of air between the probe and the surface prevents the sound signal from being transmitted, thus preventing the acquisition of the measurement\footnote{https://www.flyability.com/blog/ultrasonic-testing}. Therefore, a good coupling is needed, to which the following contribute: (1) the use of couplant, (2) the presence of magnets, (3) an appropriately-shaped probe, and (4) surface preparation. The reference force can be minimal when there is a good magnetic coupling, or when the magnetic coupling is not needed at all, as is the case for Pulse Eddy Current sensors~\cite{meng2025study}.

To conclude the overview of system design, \tabref{tab:params} summarizes the modules of the control architecture with the tuning of the corresponding parameters, which we found to be most effective through experimental validation.

\begin{table} \caption{Modules and corresponding parameters} \centering {\footnotesize 
\begin{tabular}{|c|c|c|} 
\hline Module & Parameter & Value \\ 
\hline \multirow{2}{*}{Trajectory Generation} & \texttt{max\_vel} & $0.5\,m/s$ \\ & \texttt{max\_acc} & $0.25\,m/s^2$\\ 
\hline \multirow{3}{*}{Admittance Filter} & $\mathbf{M}$ & $0.5\,\bold{I}_{3\times3}$\\ & $\mathbf{D}$ & $7.5\,\bold{I}_{3\times3}$ \\ 
& $\mathbf{K}$ & $\mathrm{diag}(30, 30, 100)$\\ 
\hline Force Observer & $\mathbf{L}$ & $7.5\,\bold{I}_{3\times3}$\\ 
\hline \multirow{3}{*}{PD Controller} & $\mathbf{K}_p$ & $\mathrm{diag}(14, 14, 20)$ \\ 
& $\mathbf{K}_v$ & $\mathrm{diag}(4, 4, 8)$\\ 
& $\mathbf{K}_{\psi}$ & 3\\ 
\hline LowPassFilter & $\omega_c$ & 31.42 rad/s \\ \hline \end{tabular} } \label{tab:params} \end{table}

%%%%%%%%%%%%%%%%%%%%%%%%%%%%%%%%%%%%%%%%%%%%%%%%%%%%%%%%%%%%%%%%%%%%%%%%%%%%%%%%
\section{Experimental Evaluation}
\label{sec:exp}

The goal of this work is to demonstrate reliable UT measurements in a challenging, unstructured environment, where a point can be approached and inspected with centimeter-level accuracy.

\begin{figure}[t]
    \centering
    \begin{subfigure}[t]{0.45\textwidth}
        \centering
        \includegraphics[width=\linewidth]{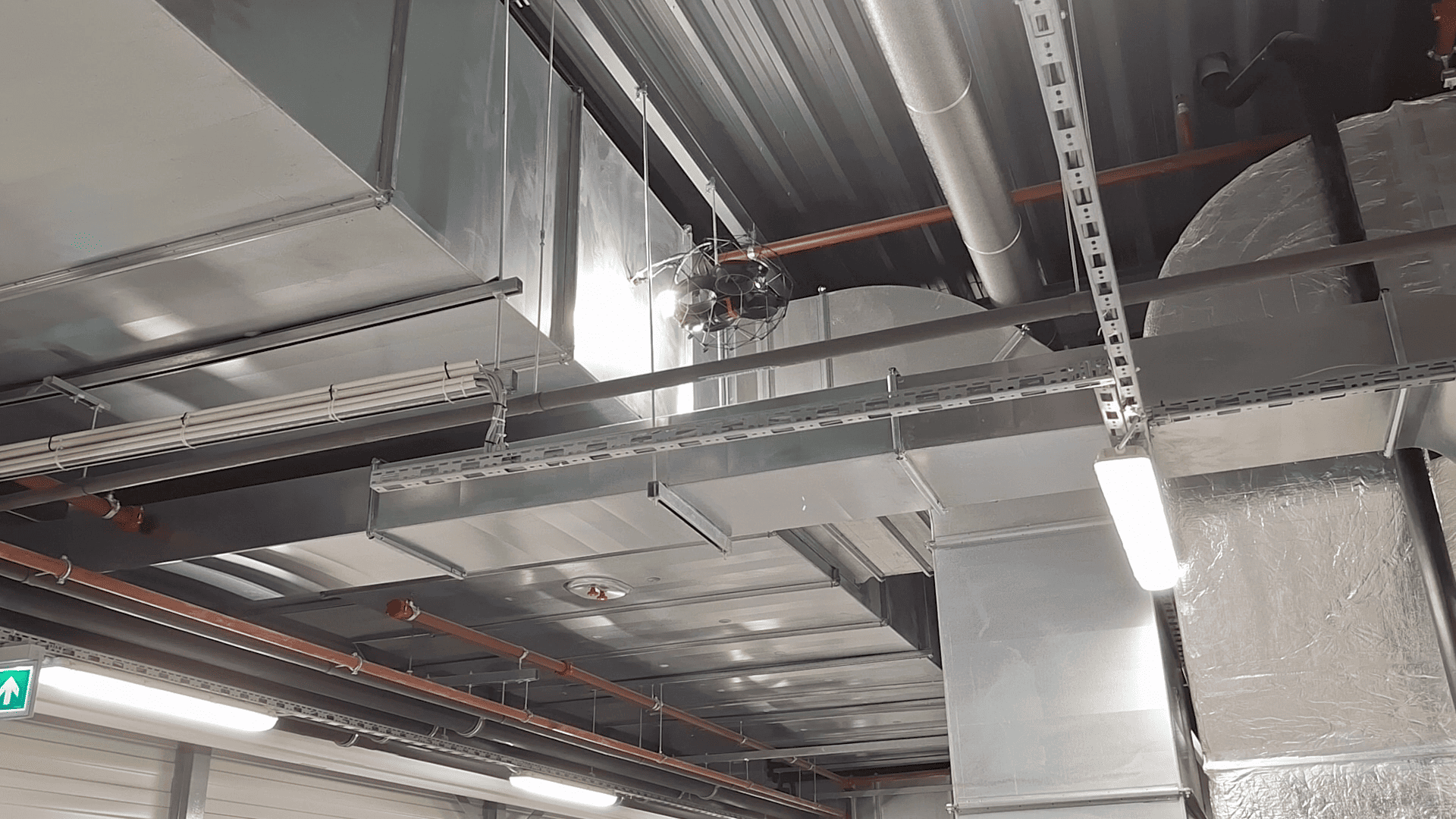}
        \caption{Wide view of the inspection location.}
    \end{subfigure}
    \begin{subfigure}[t]{0.45\textwidth}
        \centering
        \includegraphics[width=\linewidth]{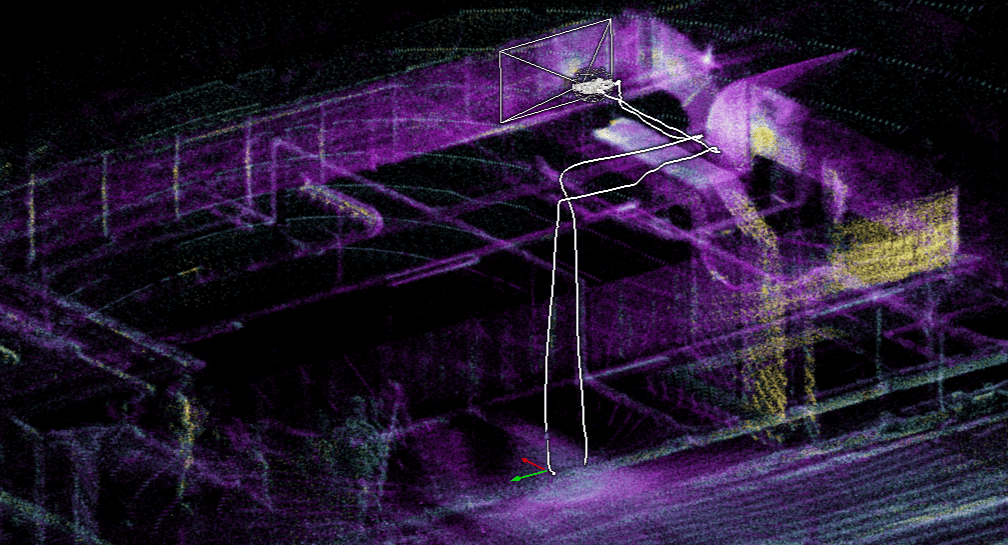}
        \caption{Top down 3D map from the Cockpit App and superimposed odometry.}
    \end{subfigure}
    \caption{(a) A wide view of the inspection location. An air duct is directly below the Elios3. The inspection takes place between rods, which may represent a collision risk. (b) The corresponding 3D map, as shown in the Cockpit App, together with the odometry.}
    \label{fig:wide}
\end{figure}

To this end, a physical inspection is conducted in a GNSS-denied industrial-like experimental drone facility available at the University of Twente, Enschede, The Netherlands. Specifically, an inspection is performed on a galvanized steel air duct in a cluttered and confined space, as shown in~\figref{fig:wide}, representing a challenging location, as such environments are expected to induce significant aerodynamic disturbances. %The pipe has a nominal wall thickness of \emph{x} [mm]. 
%For comparison a manually piloted UT inspection in the same location is also performed.

\begin{figure*}[ht]
    \centering
    \includegraphics[width=\linewidth]{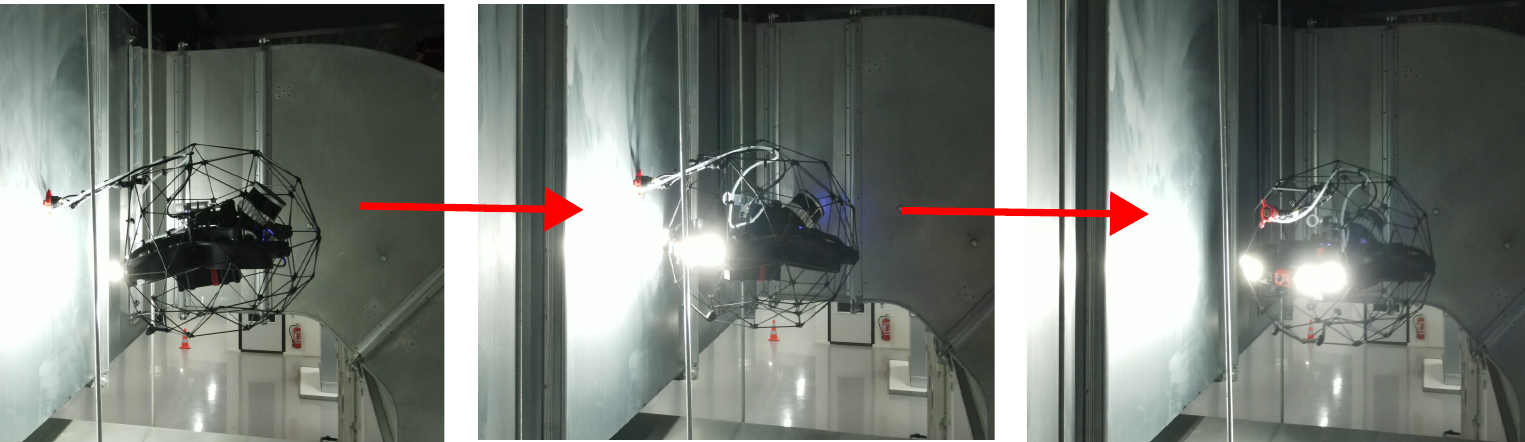}
    \caption{Detaching maneuver to disengage the magnetic probe of the Elios3 UT payload.}
    \label{fig:detach}
\end{figure*}

\begin{figure}[t]
  \centering
 \includegraphics[width=0.99\columnwidth]{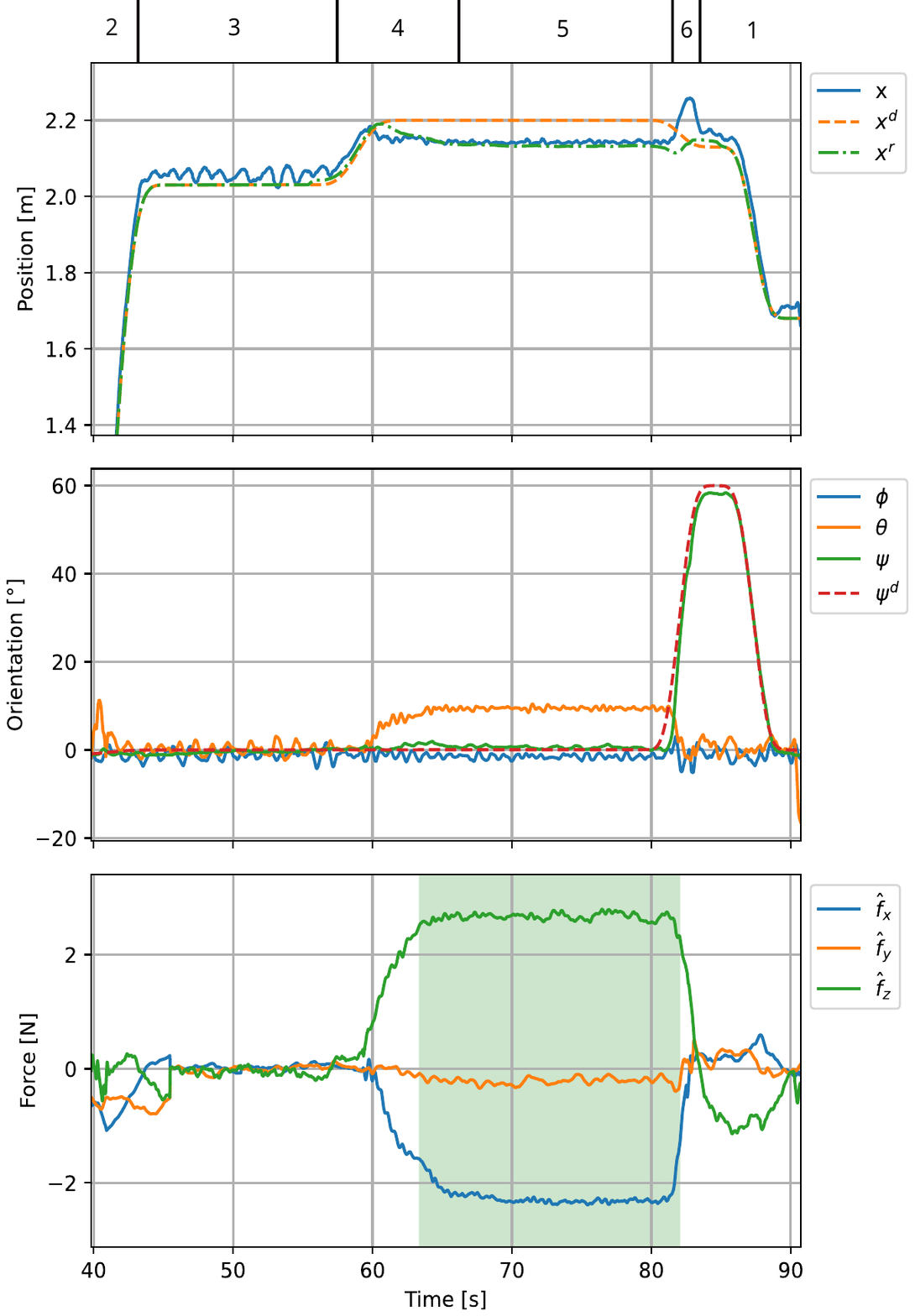}
  \caption{Position, roll-pitch-yaw ($\phi,\theta,\psi$), and force observer plots of the Elios3 during the autonomous inspection flight. All quantities are represented in the global-map/world-frame. State of the inspection is indicated above the position plot numbers correspond to states from \secref{sec:algo}. Green shading in the force plot indicate when measurement is good and stable as indicated by the Elios3 UT module.}
  \label{fig:states}
\end{figure}

%\begin{figure*}[ht!]
%    \centering
%    \begin{subfigure}[t]{0.45\textwidth}
%        \centering
%        \includegraphics[width=\linewidth]{pics/close_up_3dmap.png}
%        \caption{Close-up odometry view}
%    \end{subfigure}
%    \begin{subfigure}[t]{0.45\textwidth}
%        \centering
%        \includegraphics[width=0.7\linewidth]{pics/ut3.png}
%        \caption{UT A-Scan}
%    \end{subfigure}
%    \caption{An overview of the system operation during the contact from screenshots of the Cockpit App: (a) 4k on-board camera view; (b) close-up view of the odometry; (c) a UT A-scan, showing a \textit{good and stable} measurement.}
%    \label{fig:cockpit}
%\end{figure*}

\begin{figure}[ht]
    \centering
    \includegraphics[width=\linewidth]{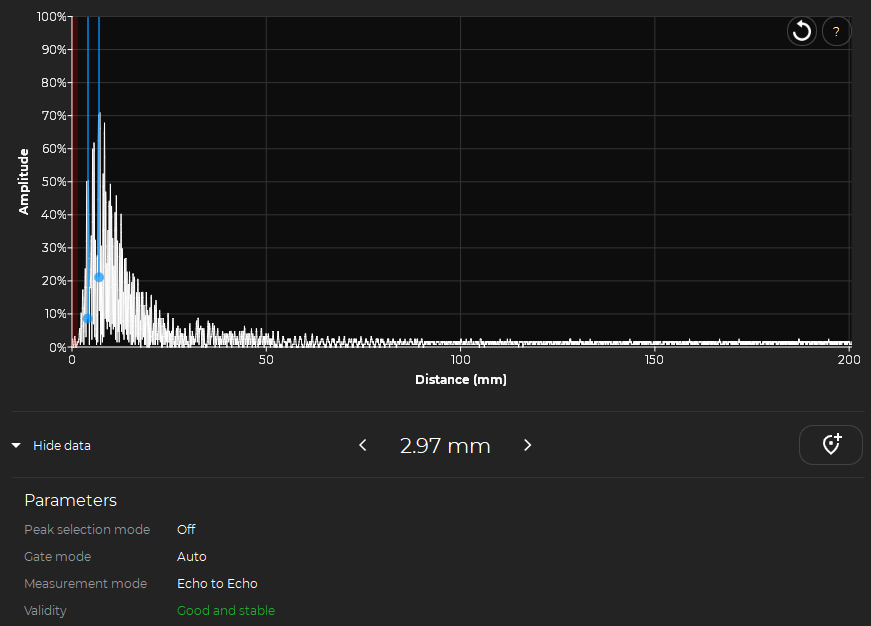}
    \caption{Screenshot of the Cockpit App of the UT A-scan, showing a \textit{good and stable} measurement.}
    \label{fig:cockpit}
\end{figure}

As discussed in \secref{sec:system-overview}, the system is first navigated to a position near the inspection point, in this case approximately 1-2 m from the target point. The desired inspection location is then provided to the inspection algorithm, which executes the measurement procedure. 

%During contact, the system must exhibit stable behavior without oscillations and apply a consistent contact force to ensure reliable measurements. While interacting with the surface, a compliant trajectory must be generated and tracked accurately to prevent excessive tracking errors that could lead to actuator saturation. Finally, the resulting UT measurements must be of high quality, as indicated by the Elios3 cockpit software, and remain consistent, without intermittent signal loss.

While interacting with the surface, a compliant trajectory must be generated and tracked accurately to prevent excessive tracking errors that could lead to actuator saturation. A contact is successful if the system (i) exhibits stable behavior without oscillations, while (ii) applying a consistent contact force to ensure reliable measurements, and (iii) the resulting UT measurements must be of high quality, as indicated by the Elios3 cockpit software, and remain consistent, without intermittent signal loss. In the following evaluation, we first analyze the performance of our system with respect to such metrics, then compare it with a manually piloted UT inspection in the same location.

%{\color{red} BB proposition of change: While interacting with the surface, a compliant trajectory must be generated and tracked accurately to prevent excessive tracking errors that could lead to actuator saturation. A contact is successful if the system (i) exhibits stable behavior without oscillations, while (ii) applying a consistent contact force to ensure reliable measurements, and (iii) the resulting UT measurements must be of high quality, as indicated by the Elios3 cockpit software, and remain consistent, without intermittent signal loss. In the following evaluation, we first analyze the performance of our system with respect to such metrics, then compare it with a manually piloted UT inspection in the same location.}

\subsection{Autonomous Inspections}
\paragraph{System Evaluation}

%To demonstrate repeatability of the operation the entire inspection operation is performed 3 times. 
The results of an autonomous inspection flight are presented in~\figref{fig:states}. The position data show that a compliant trajectory is generated in the \emph{x}-direction in response to the measured external contact force and is accurately tracked by the system. To achieve a desired steady-state contact force, the desired position can be computed from the virtual stiffness, $\boldsymbol{K}$, of the admittance filter. In this case, the depth of the trajectory is tuned to achieve a contact force of 2 N. 

The depth of the desired position to achieve a desired steady-state contact force using the virtual stiffness. 

%The RMS trajectory tracking error during the inspection procedure of each flight is summarized in ~\tabref{tab:errors}.

The attitude plot shows a slight pitch motion of the platform during interaction, which is expected as the system applies a normal force to maintain contact with the inspection surface. 

The measured contact forces are stable and consistent, exhibiting no excessive noise or spiking. Qualitative assessment from the accompanying video in the supplementary material\footnote{\url{https://codeberg.org/r-franchi/autonomous-aerial-ntd/src/branch/main/autonomousNDT_video.mp4}} further confirms the absence of oscillatory behavior during surface interaction. A force component is also observed in the \emph{z}-direction. This effect was repeatedly observed during testing and is attributed to magnetic coupling of the probe; as the probe mechanism compresses, it induces also an upward force on the cage.

%During the detachment maneuver, a transient spike in position error is observed. This is caused by the strong magnetic coupling between the probe and the surface, which requires a prying {\color{red}AF: pyring?} {\color{blue} RV: made more precise} motion to disengage. Once the magnetic contact is released, the system converges back to the desired trajectory in approximately 3 seconds. 

During the detachment maneuver, a transient spike in position error is observed. This is caused by the strong magnetic coupling between the probe and the surface, which is overcome by applying a pulling force and a yaw-induced torque on the probe to disengage the magnets. The detaching maneuver is shown in ~\figref{fig:detach}. Once the magnetic contact is released, the system converges back to the desired trajectory in approximately 1 second. An overview of the corresponding A-scan measurement, obtained from the Flyability Cockpit App, is shown in~\figref{fig:cockpit}. The interface indicates \emph{good and stable} UT measurements. The  \emph{good and stable} measurement is also indicated in the shaded region in the force plot. The measured wall thickness across all three inspections is $3\pm 0.04$ mm.

\paragraph{Repeatability Analysis}
\begin{table}[t]
\centering
\caption{RMSE errors during inspection procedure of all the autonomous flights}
\label{tab:errors}
\footnotesize
\begin{tabular}{|c|ccc|cc|}
\hline
\multirow{2}{*}{Flight} & \multicolumn{3}{c|}{RMSE [m]} & \multicolumn{2}{c|}{Position error [m]} \\
\cline{2-6}
 & $e_x$ & $e_y$ & $e_z$ & $e_y$ & $e_z$ \\
\hline
Flight 1 & 0.030 & 0.026 & 0.051 & 0.046 & 0.038 \\
Flight 2 & 0.031 & 0.025 & 0.045 & 0.052 & 0.028\\
Flight 3 & 0.031 & 0.026 & 0.046 & 0.053 & 0.027 \\
\hline
\end{tabular}
\end{table}
To demonstrate repeatability of the system, the entire inspection flight is performed 3 times.

The Root-mean-square (RMS) trajectory tracking error ($\boldsymbol{p}^r_B-\boldsymbol{p}_B$) of the entire inspection procedure are reported in ~\tabref{tab:errors}, which gives insight in the average trajectory tracking performance. 

As contact occurs along the \emph{x}-axis, the task requires precise positioning at a target point in the \emph{yz}-plane. Accordingly, the \emph{yz}-plane position error between the desired inspection point and the odometry estimate ($\boldsymbol{p}^d_B-\boldsymbol{p}_B$) during contact is reported for each flight.

\subsection{Manual inspections}
\begin{figure}[t]
  \centering
 \includegraphics[width=0.99\columnwidth]{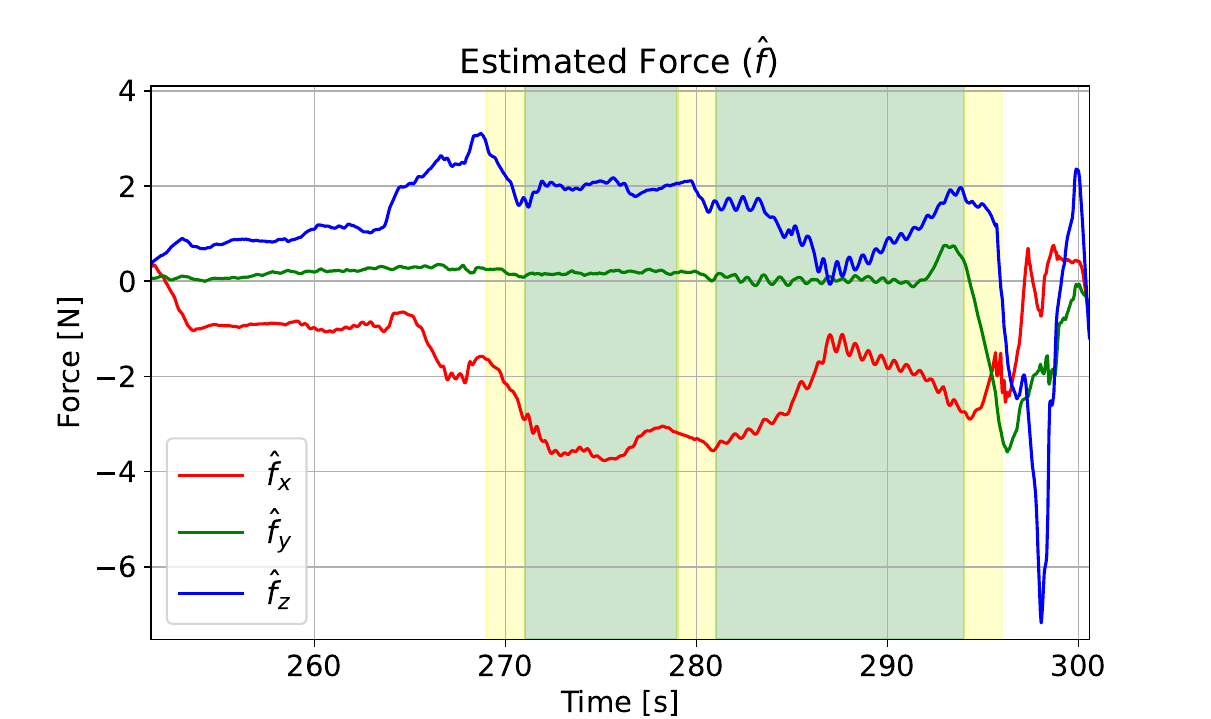}
  \caption{Force observer during manual inspections. Green and yellow shading in the force plot indicate when measurement is good and stable and unstable respectively as indicated by the Elios3 UT payload.}
  \label{fig:manual}
\end{figure}

For comparison a manually piloted UT inspection in the same location is also performed. The manual inspection is performed by an intermediate-level pilot. During this flight, measurement stability was intermittently lost, as indicated by the shaded regions in the force observer plot in ~\figref{fig:manual}. As stated earlier, there must be enough contact pressure to ensure good contact between the transducer and surface (via the couplant), however excessive pressure can squeeze the couplant from under the transducer.

\subsection{Short Discussion and Comparison}
During manual flight, it is difficult for the pilot to correct unstable measurements without substantial training and experience, as the only cues for contact force are indirect, such as hearing changes in propeller speed or vehicle orientation (e.g., pitch angle). In contrast, the autonomous inspections demonstrated high repeatability.

Overall, the results demonstrate that centimeter-level inspection accuracy (with respect to the odometry) can be achieved, thanks to a stiffly tuned controller and low tracking error, in a complex and cluttered environment, combined with an admittance control based on the contact force  estimated using the quadrotor model, the propeller speed measurements, and a relatively easy-to-identify  thrust coefficient. The system successfully combines mapping, navigation, compliant interaction, and ultrasonic measurement while maintaining stable contact and consistent measurement quality.

%%%%%%%%%%%%%%%%%%%%%%%%%%%%%%%%%%%%%%%%%%%%%%%%%%%%%%%%%%%%%%%%%%%%%%%%%%%%%%%%
\section{Conclusion}
\label{sec:conclusion}

In this work, we introduced a control paradigm for NDT contact-based aerial robotic inspection, based on the admittance filter for compliant physical interaction that can be virtually ideally used on many commercial platform conceived for GNSS-denied inspection. To the best of our knowledge, this is the first application in which the implemented control scheme interacts with the off-the-shelf low-level controller on a commercial platform.

Our control scheme and software architecture  has been customized for the onboard sensing suite of the Flyability Elios3 drone for confined-space inspection to (1)~estimate the external wrench; (2)~compute compliant and smooth trajectories; (3)~collect UT measurements of metal thickness, both in the controlled lab scenario, where the early testing occurred, and in the industrial-like facility at the Univeristy of Twente, which we show in this paper. Our method exploits the aerial physical interaction paradigm introduced by some of our previous works and other works in the literature of redundant aerial manipulators for NDT inspection, and the onboard sensing modalities and interfaces provided by the commercial platform.

We show the performances of all the constitutive components of the proposed architecture: (1) the force estimator; (2) the trajectory following; (3) the achieved UT measurements. Despite these encouraging results, there is space for improvement. For example, we plan on (i) further automating the procedure using visual servoing algorithms; (ii) characterizing the connection between force estimate and quality of the NDT measurement; (iii)  evaluate the measurement performance of the interaction scheme and the associated measurement quality for non-ferromagnetic materials, for which magnetic coupling is reduced.

%%%%%%%%%%%%%%%%%%%%%%%%%%%%%%%%%%%%%%%%%%%%%%%%%%%%%%%%%%%%%%%%%%%%%%%%%%%%%%%%
\section{Acknowledgments}
\label{sec:ack}

The authors thank: the Flyability team collaborating in the AutoASSESS project for the platform and the software interface; the Campus Facility Management of the University of Twente for making available industrial-like facilities for field tests; Quentin Sablé and Arthur Gasienica for their constant support in the lab; the partners and the coordinator of the AUTOASSESS project for the inspiring and constructive technical exchanges during consortium meetings.

%%%%%%%%%%%%%%%%%%%%%%%%%%%%%%%%%%%%%%%%%%%%%%%%%%%%%%%%%%%%%%%%%%%%%%%%%%%%%%%%
% If applicable
%\section*{Acknowledgments}
%We thank XXX for fruitful discussions and for ...

\printbibliography[title={References}]

\end{document}